\documentclass[letterpaper,10pt]{article}
\usepackage{cite}
\usepackage{amsmath,amssymb,amsfonts}
\usepackage{algorithmic}
\usepackage{graphicx}
\usepackage{textcomp}
\usepackage{xcolor}
\usepackage{wrapfig}
\usepackage{caption}
\usepackage{subcaption}
\usepackage{titling}
\def\BibTeX{{\rm B\kern-.05em{\sc i\kern-.025em b}\kern-.08em
    T\kern-.1667em\lower.7ex\hbox{E}\kern-.125emX}}
\begin{document}

\title{Facial Emotion Recognition: A multi-task approach using deep learning}

\author{{1\textsuperscript{st},* Aakash Saroop}\\
{\textit{Department of Computer Engineering}} \\
{\textit{K. J. Somaiya College of Engineering}}\\
{Mumbai, Maharashtra}\\
{aakash.saroop@somaiya.edu}\\
{*Corresponding author}\\
\and
{2\textsuperscript{nd} Pathik Ghugare}\\
{\textit{Department of Computer Engineering}}\\
{\textit{K. J. Somaiya College of Engineering}}\\
{Mumbai, Maharashtra} \\
{pathik.g@somaiya.edu}\\
\and
{3\textsuperscript{rd} Sashank Mathamsetty}\\
{\textit{Department of Computer Engineering}}\\
{\textit{K. J. Somaiya College of Engineering}}\\
{Mumbai, Maharashtra} \\
{sashank.m@somaiya.edu }\\
\and
{4\textsuperscript{th} Vaibhav Vasani}\\
{\textit{Department of Computer Engineering}}\\ 
{\textit{K. J. Somaiya College of Engineering}}\\
{Mumbai, Maharashtra} \\
{vaibhav.vasani@somaiya.edu}\\
}

\maketitle

\begin{abstract}
Facial Emotion Recognition is an inherently difficult problem, due to vast differences in facial structures of individuals and ambiguity in the emotion displayed by a person. Recently, a lot of work is being done in the field of Facial Emotion Recognition, and the performance of the CNNs for this task has been inferior compared to the results achieved by CNNs in other fields like Object detection, Facial recognition etc. In this paper, we propose a multi-task learning algorithm, in which a single CNN detects gender, age and race of the subject along with their emotion. We validate this proposed methodology using two datasets containing real-world images. The results show that this approach is significantly better than the current State of the art algorithms for this task.

\end{abstract}

Keywords- Facial Emotion Recognition, Multi-task learning, Multi-output model, Convolutional Neural Networks, Deep Learning, Image Processing

\section{Introduction}
As AI continues to become an increasing integral part of our lives, the need for
machines to understand the state of the human emotions is of critical importance. This has the potential to take human computer interaction to the next level, with a direct impact in the field of voice assistants, mental health therapy, recommendation systems etc.

Since there are unique local languages and local moral orders, different cultures can use the same emotion and expression in very different ways [1]. This makes facial emotion recognition is an ambiguous task as each person picks up visual cues differently. The State of the art algorithms for Facial Emotion Recognition (FER) give much lower accuracy on a reasonably large dataset obtained from real world images taken in a non controlled environment [2, 17], as compared to other computer vision tasks like object detection [18], image classification [19] etc.

In this paper, we hypothesise that a Convolutional Neural Network, subjected to multi- task learning on the same data, i.e. learning to predict emotion, age, gender and race of a subject from the same facial image would perform better on individual tasks, by transferring knowledge across different domains.
The summary of contributions through this paper is:
\begin{itemize}
    \item Validation that the proposed multi-task learning approach of combining FER with age, gender and race classification outperforms the multi-task approach of combining FER and facial action unit detection and the conventional single-task learning approach.
    \item Presenting a CNN architecture which gives outputs corresponding to all the labels mentioned above.
\end{itemize}

The rest of this paper is organised as follows: Related work has been reviewed in Section II. The proposed approach has been discussed in Section III. Experimental  setup has been shared in Section IV. Experimental results are shared in Section V. Further discussions on the observations made are discussed in Section VI. Conclusion and future work are suggested in Section VII.

\section{Related works}

The work done in the field of FER can be broadly divided into two categories based on whether the features were hand-crafted or generate through a Neural Network.
Before the widespread use of CNNs in the field of computer vision, FER used to be carried out by identifying facial components or landmarks from the facial region [23, 24, 25, 26].
Recent work in the field of Facial Emotion Recognition deals with using Real World Images instead of taking images in a controlled environment. This leads to several pre-processing methods being used on the images obtained . 
Some of these approaches are used for preprocessing the images, before feeding them to a CNN[2, 5]. In [2], feature extraction approaches such as HoG, LBP are used to extract features from the image, which requires relatively lower computing power and memory than Deep Learning approaches. In [5], Milad Zadeh et. al. discusses the use of pre-processing method of Gabor filters to increase the accuracy of Facial emotion recognition. In this paper,  a very small dataset [12] of 213 images which lead to overfitting on the Neural Network. [14] draws the Bezier curve on the eye and mouth and classifies the emotion of the characteristic with Hausdroff distance. 

Ever since the in popularity of Deep Learning, CNNs have been used for FER, which provide the output by enabling “end-to-end” learning to occur in the 
pipeline directly from the input images [27].
In [4], Mehdipour Ghazi et. al. discusses an approach to deal with Emotion recognition using DL under several conditions including the varying head pose angles, upper and lower face occlusion, changing illumination of different strengths, and misalignment due to erroneous facial feature localisation. They have proposed that using preprocessing methods for pose and illumination normalisation along with pre-trained deep learning models or accounting for these variations during training substantially resolve this weakness present in the  dataset.
[15] approaches the problem of emotion classification based on thermal images of the face. \\
The ideas presented in this paper have been heavily built upon the work done in [3]. Gerard Pons et. al. take the work of multi-task, multi-label and multi-domain learning with residual convolutional networks for emotion recognition. They had claimed that using the same CNN, when trained to predict both emotion and facial action units, it performed better at each individual task. In this paper, it was suggested that other features.

\section{Proposed Aprroach}
Here the approach of multi-task learning, i.e. a single model generating labels for emotion, gender, age and race, is to be compared with the approach of single-task learning, i.e. a model predicting only emotion of a subject, as shown in Figure1.
The CNN architecture used for this study is shown in Figure2. More details about the CNN can be found in the appendix.

\begin{figure}[htbp]
\centerline{\includegraphics[width=1\textwidth]{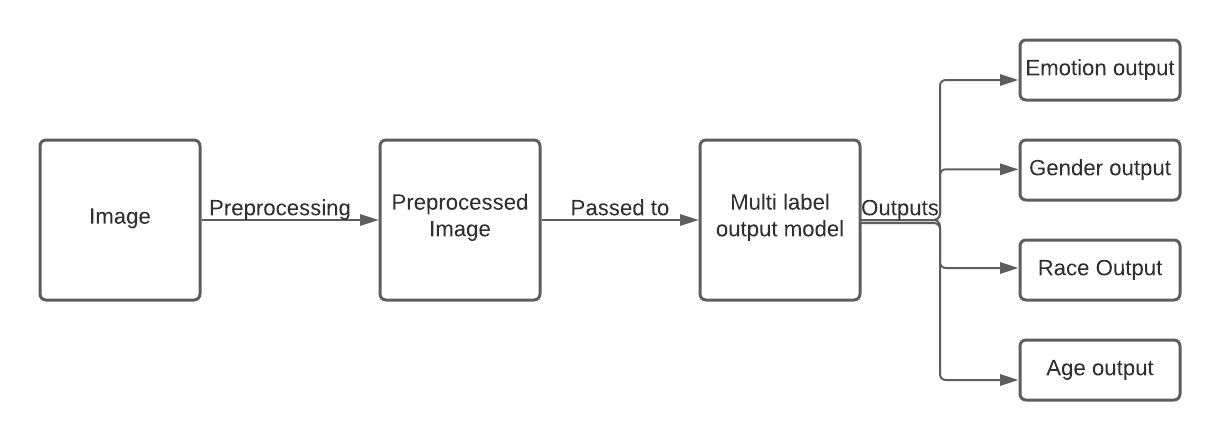}}
\caption{Architecture of the model}
\label{fig}
\end{figure}

Each of the classification tasks have a separate output layer, which means that 
each output layer has a softmax function which is independent from the softmax function of other output layers:

\begin{equation}
\sigma(z)_i = \frac{e^{z_i}}{\sum_{j=1}^{K}{e^{z_j}}} 
\end{equation}
\\
$\sigma=$ softmax\\
$K=$ number of classes of the classifier\\
$e^{z_i}=$ standard exponential function for  vector\\
$e^{z_j}=$ standard exponential function for  vector\\

For emotion recognition, the model classifies the input image in one of 7 classes consisting of the basic human emotions: Surprise, fear, disgust, happy, sad, angry and neutral. Gender is classified as male, female or unsure. Race is classified as Caucasian, African-American or Asian. For age estimation, the classes are divided into a range of ages: 0-3, 4-19, 20-39, 40-69 and  70+. The datasets used are FER, containing labels for only emotion and RAF-Db containing labels corresponding to all 4 tasks.

\begin{figure}[htbp]
\centerline{\includegraphics[width=1\textwidth]{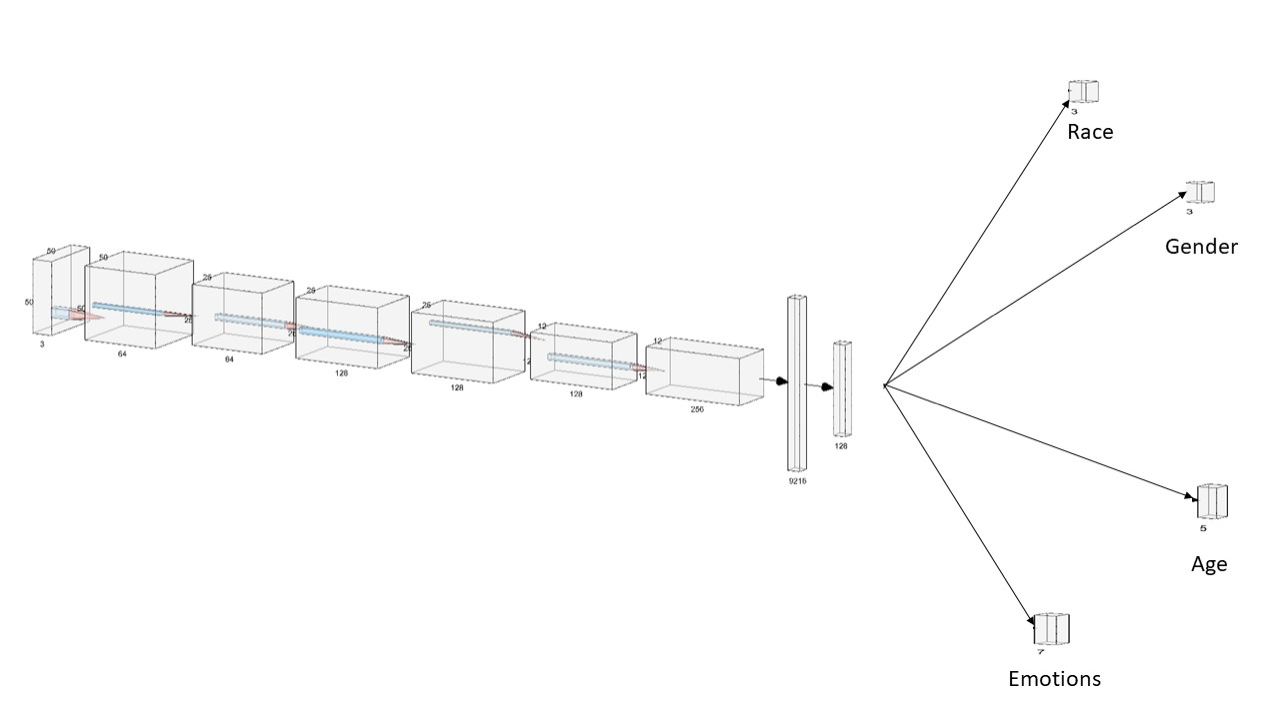}}
\caption{Architecture of the model}
\label{fig}
\end{figure}

In the pre-processing stage, the images have been subjected to pose normalisation. This has been achieved by identifying the eye centres of the faces presented in the images, and the images being rotated such that the line joining the eye centres becomes horizontal. The eye detection has been carried out using the Cascade classifier of opencv library [16]. There were a few cases in which the pre-trained algorithm was incorrectly identifying the eye centres and the images were being rotated by an angle of large magnitude, distorting the images(Figure 3). To counter this, the max range to which the image can be rotated has been capped at 10 degrees. In the images presented in both the data sets, the faces only have minor rotation presented. Capping the maximum rotation at 10 degrees ensures that the examples in which eye centers are identified correctly are normalised, but the examples in which eyes are not correctly identified, are not rotated by a large magnitude.

The CNN architecture has been strongly built upon [6], with the change in dropout rates to improve the accuracy.
The dropout rates in [6] were 0.4, 0.4, 0.5, 0.6 in increasing layers of  the neural network. In the neural network architecture presented in this paper, the dropout layers are now monotonically decreasing : 0.6, 0.5, 0.4, 0.4 in the increasing layers of the neural network.
\begin{figure}
     \centering
     \begin{subfigure}[b]{0.3\textwidth}
         \centering
         \includegraphics[width=\textwidth]{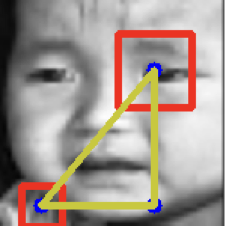}
         
         \caption{}
     \end{subfigure}
     \begin{subfigure}[b]{0.3\textwidth}
         \centering
         \includegraphics[width=\textwidth]{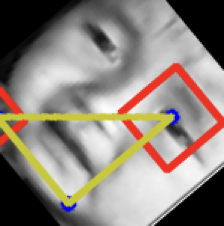}
       \caption{}
     \end{subfigure}
        \caption{(a) Showing the original image in the dataset with incorrectly identified eye centers. (b) Showing the distortion to the pre-processed image
}
        
\end{figure}

On the FER dataset, the cross validation accuracy of the previous model was 63\%.
Upon making the changes, the accuracy has increased to 67\%.
The reasoning for why this change having performed better than the original model could be: in deeper layers of the Neural Network, each layer learns to detect more complex patterns [7], which are of increasing importance in our classification task, and need to  have higher probability of being retained in each pass through the neural  network.
The second last layer of the neural network provides an input to 4 separate output layers, one for each of the classification tasks.
So, except the output layer, all the classification tasks have the same weights for the rest of the neural network.
For each of the classification tasks, categorical cross entropy is used to calculate the loss of each task. Hence there are  4 loss functions. To calculate the loss function of the entire Neural network, adequate weights needs to be assigned to the loss function of each of the tasks, to get a meaningful result in each of the separate tasks. The weights assigned to emotion, age, race and gender loss were 2, 4, 1.5, 0.1 respectively.

Using the above weights, the overall loss is calculated. Then the model backpropagates this loss to improve its performance in subsequent iterations.
To ensure that the CNN gives optimal performance, two callbacks were used:

Early stopping: The validation accuracy for emotion was calculated for each epoch, during the training of the neural network. This is used to determine when the model started overfitting on the data. When overfitting has been identified, further training of the model is stopped and the weights from the epoch which provided the best validation accuracy for the data, have been restored to the final model.

Reduce on Plateau: If the validation accuracy for emotion output fails to improve for 5 epochs, the learning rate is reduced to 0.2 times the previous model to continue training of the model.

\section{Experimental Setup}
In this paper the following datasets have been used:
RAF-DB [9]: Real-world Affective Faces Database (RAF-DB) is a large-scale facial expression database with around 30K great-diverse facial images downloaded from the Internet. Based on the crowdsourcing annotation, each image has been independently labeled by about 40 annotators. Images in this database are of great variability in subjects' age, gender and ethnicity, head poses, lighting conditions, occlusions, (e.g. glasses, facial hair or self-occlusion), post-processing operations (e.g. various filters and special effects), etc. RAF-DB has large diversities, large quantities, and rich annotations.\\
FER [8]: The FER dataset classifies facial expressions from 35,685 examples of 48x48 pixel grayscale images of faces. Images are categorized based on the emotion shown in the facial expressions (happiness, neutral, sadness, anger, surprise, disgust, fear).

The model was given the data in the form of \{$x_i,y_i$\}, where $x_i$ is a 50x50 grayscale image containing a human face, and $y_i$ is the list of numpy arrays containing the labels corresponding to the image.
A sample output i:\\
$[[1, 0, 0, 0, 0, 0, 0], [1, 0, 0], [1, 0, 0],[1, 0, 0, 0, 0]]$\\
Signifies that the person present in the image is surprised, male, caucasian and has an age of 0-3.

The multi label CNN was trained using a batch size of 32 images.
The initial learning rate was set to 3e-4 and the number of epochs was equal to 100.
All the models were implemented in Tensorflow [13].

\section{Experimental results}
The validation accuracies recorded can be seen in Table 1.
\begin{table}[htbp]

\begin{center}
\begin{tabular}{|c|c|c|}

\hline

\textbf{Dataset} & \textbf{\textit{Single Label 
Output (Emotion)}}& \textbf{\textit{Multi Label 
Output}} \\
\hline
\textbf{RAFDb} & \textbf{\textit{0.4538}}& \textbf{\textit{0.7926}} \\
\hline
\textbf{FER} & \textbf{\textit{0.6791}}& \textbf{\textit{0.5312*}} \\
\hline

\multicolumn{3}{l}{$^{\mathrm{*}}$FER has labels only corresponding to emotions. 
}

\end{tabular}
\label{tab1}
\end{center}
\caption{Emotion Validation Accuracies}
\end{table}
In the multi label output model, the validation accuracy of the model after being trained on RAFDB is 79.26\%. After the same model is further trained on FER, its accuracy reduces to 53.12\%. Hence, for our final model, the training was stopped after just RAFDB and the final accuracy is 79.26\%. This is an improvement from the accuracy reported in the multi-task model of [3], which gives an accuracy of 45.9\% for emotion classification on the SFEW dataset. This shows that the hypothesis of this paper is correct.

\begin{table}[htbp]

\begin{center}
\begin{tabular}{|c|c|c|}

\hline

\textbf{Label} & \textbf{\textit{RAFDB
 (Basic Emotions)}}& \textbf{\textit{FER (After training on RAFDB)}} \\
\hline
\textbf{emotion} & \textbf{\textit{0.7926}}& \textbf{\textit{0.5312}} \\
\hline
\textbf{gender} & \textbf{\textit{0.7832}}& \textbf{\textit{N/A*}} \\
\hline
\textbf{race/ ethnicity} & \textbf{\textit{0.8610 }}& \textbf{\textit{N/A*}} \\
\hline
\textbf{age} & \textbf{\textit{0.7476 }}& \textbf{\textit{N/A*}} \\
\hline
\multicolumn{3}{l}{$^{\mathrm{*}}$Not applicable. In the Multi-label model, weights for other losses set to 0.
}

\end{tabular}
\label{tab1}
\end{center}
\caption{Cross Validation Accuracies}
\end{table}
For each of the datasets, 90\% of the images were used as training data and 10\% of the images were used as cross-validation data. As seen in Figure 4, as the accuracy of the model begins to plateau multiple times throughout the training, the learning rate of the model is reduced before further training. However, when the learning rate becomes less than 1e-6, further training is stopped and the model weights from the epoch having the best validation accuracy are restored to the final model.
\begin{table}[htbp]

\begin{center}
\begin{tabular}{|c|c|c|}

\hline

\textbf{Label} & \textbf{\textit{RAFDB
 (Basic Emotions)}}& \textbf{\textit{FER (After training on RAFDB)}} \\
\hline
\textbf{emotion} & \textbf{\textit{0.6956}}& \textbf{\textit{1.2371}} \\
\hline
\textbf{gender} & \textbf{\textit{0.5110}}& \textbf{\textit{N/A*}} \\
\hline
\textbf{race/ ethnicity} & \textbf{\textit{0.4524}}& \textbf{\textit{N/A*}} \\
\hline
\textbf{age} & \textbf{\textit{0.9003}}& \textbf{\textit{N/A*}} \\
\hline
\multicolumn{3}{l}{$^{\mathrm{*}}$Not applicable. In the Multi-label model, weights for other losses set to 0.
}

\end{tabular}
\label{tab1e}
\end{center}
\caption{Loss values}
\end{table}

\begin{figure}
     \centering
     \begin{subfigure}[b]{0.5\textwidth}
         \centering
         \includegraphics[width=\textwidth]{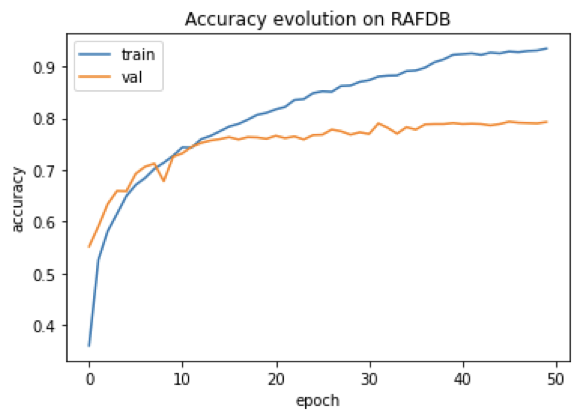}

     \end{subfigure}
     \begin{subfigure}[b]{0.5\textwidth}
         \centering
         \includegraphics[width=\textwidth]{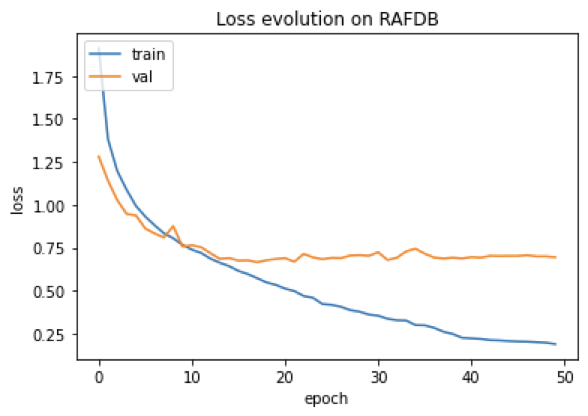}
         
       \caption{} 
     \end{subfigure}
     \begin{subfigure}[b]{0.5\textwidth}
         \centering
         \includegraphics[width=\textwidth]{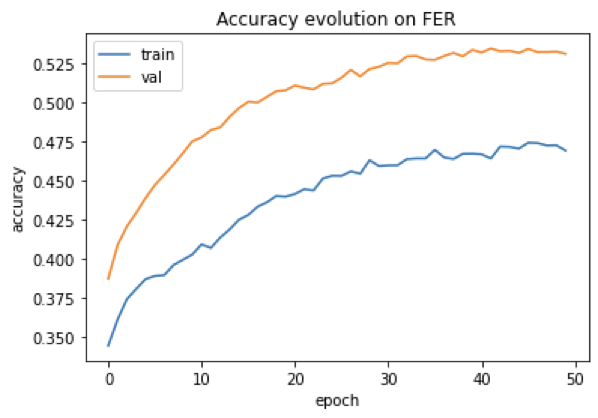}

     \end{subfigure}
     \begin{subfigure}[b]{0.5\textwidth}
         \centering
         \includegraphics[width=\textwidth]{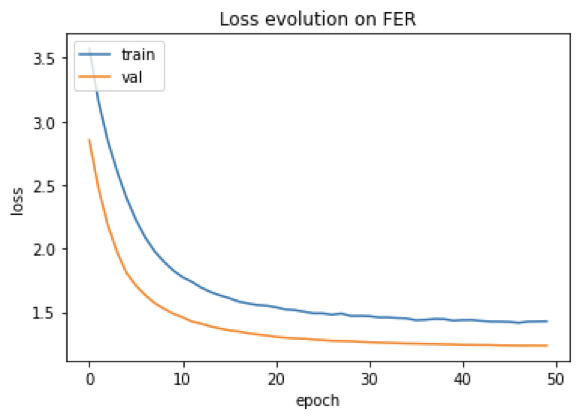}
         
        \caption{}
     \end{subfigure}
        \caption{ Accuracy and loss evolution of the label emotion.(a) On RAF-Db (b) On FER after training on RAF-Db}

\end{figure}
Some parts of the code were taken from [6, 10, 11].\\
In addition to this, while trying to perform transfer learning on neural networks which had been pre-trained on facial recognition, the following accuracies were obtained for single task learning of facial emotions on the FER dataset:\\
VGG16 [20]: 18.03\%\\
ResNet50v2 [21]: 48.08\%\\
FaceNet [22]: 30.88\%\\
These are all lower than the accuracy obtained on the neural network used for single task learning without any transfer learning, which was 63\%.Hence the final model used did not use transfer learning.
Sample outputs of the model are shown in Figure 5.

\section{Discussions}
In case of multi-task learning the validation accuracy is 79\% when trained for all the labels on RAF-Db, which reduces to 53\% when the same model, pre-trained on RAF-Db was further trained single task learning on FER, as this dataset has labels only corresponding to emotions.
Hence the final multi-task learning model, which gives the best results, has been trained only on RAF-Db.

\section{Conclusion and Future work}
In this paper we developed an approach to perform multi-task learning, i.e. predicting gender, age and race along with emotion. The results obtained were drastically better than the classical approach of single task learning using the same CNN architecture. This paper was an effort at improving the accuracy of CNNs for Facial Emotion  Recognition.
Future work will consist of: 
\begin{itemize}
    \item The approach of multi-task learning can be tested on various CNN architectures for facial emotion recognition.
    \item To effectively train more data having all the labels. Labels for age, gender and race can be generated using pre-trained open source models for the FER dataset and used as a part of the training set.
    \item Various filters such as Gabor, HOG, LBP, SIFT can be applied during the preprocessing step and their effect on the results can be studied.
    \item Finding the optimal values of the weights of the loss function of the 4 branches of the CNN, instead of hardcoding the values.
    \item Finding the optimal values of the weights of the loss function of the 4 branches of the CNN, instead of hardcoding the values.

\end{itemize}

\begin{figure}
     \centering
     \begin{subfigure}[b]{0.15\textwidth}
         \centering
         \includegraphics[width=\textwidth]{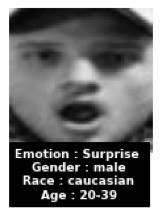}

     \end{subfigure}
     \begin{subfigure}[b]{0.15\textwidth}
         \centering
         \includegraphics[width=\textwidth]{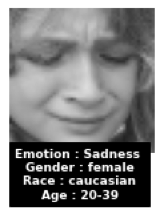}

     \end{subfigure}
     \begin{subfigure}[b]{0.15\textwidth}
         \centering
         \includegraphics[width=\textwidth]{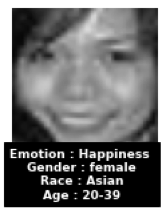}

     \end{subfigure}
     \begin{subfigure}[b]{0.15\textwidth}
         \centering
         \includegraphics[width=\textwidth]{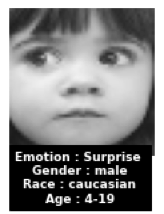}

     \end{subfigure}
     \begin{subfigure}[b]{0.15\textwidth}
         \centering
         \includegraphics[width=\textwidth]{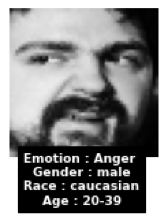}
         
     \end{subfigure}
        \caption{Sample Outputs of the model}

\end{figure}

\section*{Acknowledgment}
We would like to thank Li, Shan and Deng, Weihong and Du, JunPing for giving us access to the RAF-Db dataset. We are also grateful to K. J. Somaiya College of Engineering for giving us the opportunity to work in this research area.

\vspace{12pt}
Appendix: The appendix contains the model architecture of the CNN used
\begin{figure}[htbp]
\centerline{\includegraphics[width=1.4\textwidth]{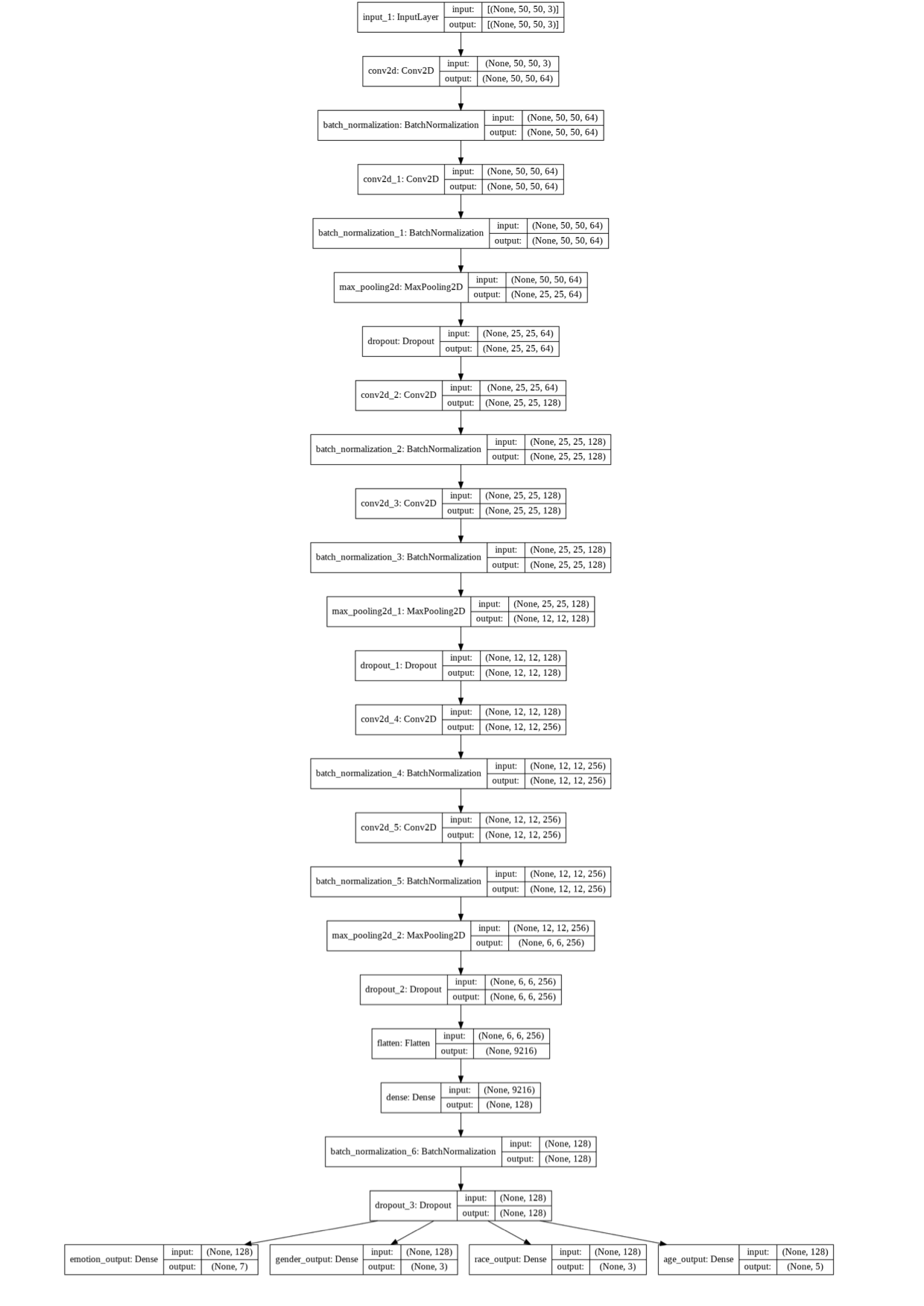}}

\label{fig}
\end{figure}

\end{document}